\documentclass{article}

\PassOptionsToPackage{numbers,sort&compress}{natbib}
\usepackage[preprint]{neurips_2026}
\usepackage{multirow}
\usepackage{graphicx}
\usepackage{caption}

\usepackage[utf8]{inputenc} 
\usepackage[T1]{fontenc}    
\usepackage{hyperref}       
\usepackage{url}            
\usepackage{booktabs}       
\usepackage{amsfonts}       
\usepackage{nicefrac}       
\usepackage{microtype}      
\usepackage{xcolor}         
\usepackage{wrapfig}
\usepackage[most]{tcolorbox}
\usepackage{listings}

\definecolor{boxblue}{RGB}{232,244,255}
\definecolor{boxblueframe}{RGB}{45,105,170}
\definecolor{boxyellow}{RGB}{255,248,218}
\definecolor{boxyellowframe}{RGB}{185,135,20}
\definecolor{boxgreen}{RGB}{235,250,235}
\definecolor{boxgreenframe}{RGB}{55,135,65}
\definecolor{boxgray}{RGB}{246,246,246}
\definecolor{boxgrayframe}{RGB}{120,120,120}

\newtcolorbox{bluecasebox}[1]{
  colback=boxblue,
  colframe=boxblueframe,
  title=\textbf{#1},
  fonttitle=\bfseries,
  arc=2mm,
  boxrule=0.8pt,
  left=1.5mm,
  right=1.5mm,
  top=1mm,
  bottom=1mm
}

\newtcolorbox{yellowcasebox}[1]{
  colback=boxyellow,
  colframe=boxyellowframe,
  title=\textbf{#1},
  fonttitle=\bfseries,
  arc=2mm,
  boxrule=0.8pt,
  left=1.5mm,
  right=1.5mm,
  top=1mm,
  bottom=1mm
}

\newtcolorbox{greencasebox}[1]{
  colback=boxgreen,
  colframe=boxgreenframe,
  title=\textbf{#1},
  fonttitle=\bfseries,
  arc=2mm,
  boxrule=0.8pt,
  left=1.5mm,
  right=1.5mm,
  top=1mm,
  bottom=1mm
}

\newtcolorbox{graycasebox}[1]{
  colback=boxgray,
  colframe=boxgrayframe,
  title=\textbf{#1},
  fonttitle=\bfseries,
  arc=2mm,
  boxrule=0.8pt,
  left=1.5mm,
  right=1.5mm,
  top=1mm,
  bottom=1mm
}

\newtcolorbox{claimbox}{
  colback=gray!10,
  colframe=gray!65,
  boxrule=0.8pt,
  arc=2mm,
  left=6pt,
  right=6pt,
  top=5pt,
  bottom=5pt,
  before skip=8pt,
  after skip=8pt
}

\title{Confidence Laundering in Agent Systems: Why Uncertainty Needs a Latent Carrier}

\author{%
  Kaiwen Shi,
  Zheyuan Zhang,
  Han Bao,
  Colby Nelson,
  Yanfang Ye$^{\dagger}$ \\
  University of Notre Dame \\
  \texttt{\{kshi3, yye7\}@nd.edu} \\
  \vspace{0.5em}
  {\small $^{\dagger}$Corresponding author}
}

\begin{document}

\maketitle

\begin{abstract}
  Modern agent systems can turn uncertainty into overconfidence: fragile upstream decisions are exposed as clean intermediate artifacts, while downstream components inherit the commitment without the uncertainty behind it. As a result, local ambiguity can become system-level error amplification. We argue that this reveals an interface bottleneck in agent uncertainty propagation: uncertainty does not propagate merely because a trajectory contains uncertain steps; \textbf{it propagates only when it survives the handoff between components.} We define \textit{uncertain decision handoff} as the transfer of an intermediate decision made under uncertainty, and identify \textit{confidence laundering} as the failure mode in which fragile upstream states are repackaged as procedurally valid artifacts that downstream agents over-trust. To address this bottleneck, we propose \textit{latent uncertainty} as an uncertainty-bearing carrier attached to decision handoffs. Rather than replacing text with hidden states, latent uncertainty aims to preserve pre-commitment fragility in a downstream-usable form. This position shifts agent uncertainty propagation from step-wise estimation toward uncertainty-preserving interface design for more recoverable agent systems.
\end{abstract}

\section{Introduction}
Modern language agents, particularly those powered by large language models (LLMs) \cite{ye2025llms4all,chen2025obvious,chen2025clear}, increasingly operate as multi-step systems that decompose complex tasks into subtasks, invoke tools, retrieve external knowledge, generate intermediate artifacts, and propagate partial results through interconnected reasoning and execution workflows \cite{zhang2026agentic,schick2023toolformer,ohuncertainty,zhang2025agentrouter}. In such systems, ignoring uncertainty propagation creates a structural source of overconfidence \cite{ohuncertainty,duan2025uprop,zhao2024saup}. Uncertain upstream decisions can be transformed into apparently stable intermediate artifacts, causing downstream components to reason, verify, and act on commitments whose fragility has been erased~\citep{zhang2023language,zhu2025llm,guthrie2005spark,zhang2025graphtracer}. The consequence is not only local miscalibration, but system-level error amplification: uncertainty that should trigger caution, clarification, recovery, or rerouting is instead hidden behind fluent text, retrieved evidence, or executable actions \cite{zhu2025llm,guthrie2005spark}. This makes uncertainty a systems-level problem rather than only a prediction-level property. Uncertainty arises when an agent selects a tool, formulates a query, interprets an observation, chooses evidence, routes a task to another agent, or commits to an intermediate result \cite{liu2024uncertainty,kuhn2023semantic,zhang2025agentrouter}. Recent work on agent uncertainty propagation moves beyond single-step uncertainty estimation by modeling how uncertainty accumulates across trajectories \cite{zhao2025uncertainty, duan2025uprop, zhao2024saup, tayebati2026tracer, zhang2026selaur}. Yet most formulations still treat propagation as trajectory-level estimation: estimating uncertainty at each step, decomposing its sources, or aggregating it over time. This leaves a basic systems question unresolved: how does uncertainty move from one component to the next?

The answer is that \textbf{uncertainty propagates through interfaces only when the fragility of an upstream decision is preserved for downstream use}. However, agent interfaces usually transmit committed artifacts, such as tool calls, queries, API arguments, retrieved passages, structured fields, routing decisions, or textual answers \cite{schick2023toolformer, lewis2020retrieval,zhang2025agentrouter}. These artifacts are useful because they are discrete and executable, but this usefulness comes at an epistemic cost: they reveal what the upstream agent decided, not how fragile that decision was \cite{lin2022teaching}. We call this setting \textit{uncertain decision handoff}: an upstream component makes an intermediate decision under uncertainty and passes the resulting artifact downstream for further reasoning or action. This is not mere communication, because the downstream component uses the artifact as a premise for computation, tool execution, evidence selection, routing, or answer generation. Losing uncertainty at the handoff can therefore change the behavior of the whole system \cite{li2024survey}. The central risk is \textit{confidence laundering}: \textbf{a fragile epistemic state is converted into a well-formed symbolic or executable object, causing downstream components to treat it as more reliable than warranted} \cite{liu2025not,shorinwa2025survey}.


This motivates our central claim: the missing object in agent uncertainty propagation is not another confidence score, but a carrier that allows doubt to survive the handoff. We propose \textit{latent uncertainty}: an uncertainty-bearing representation derived from an agent's internal continuous state and attached to a decision handoff. Unlike scalar confidence or verbal explanation, latent uncertainty is a representation-level carrier. It may be probed into a score, but its primary role is to preserve uncertainty for downstream control. This paper makes four contributions. \textbf{First}, we reframe agent uncertainty propagation as an interface problem. \textbf{Second} we define uncertain decision handoff and confidence laundering. \textbf{Third}, we introduce latent uncertainty as a carrier for pre-commitment fragility. \textbf{Fourth}, we provide validation experiments showing that committed interfaces hide upstream uncertainty and that latent uncertainty can support downstream recovery behavior.

\begin{figure*}[t]
    \centering
    \includegraphics[width=0.8\textwidth]{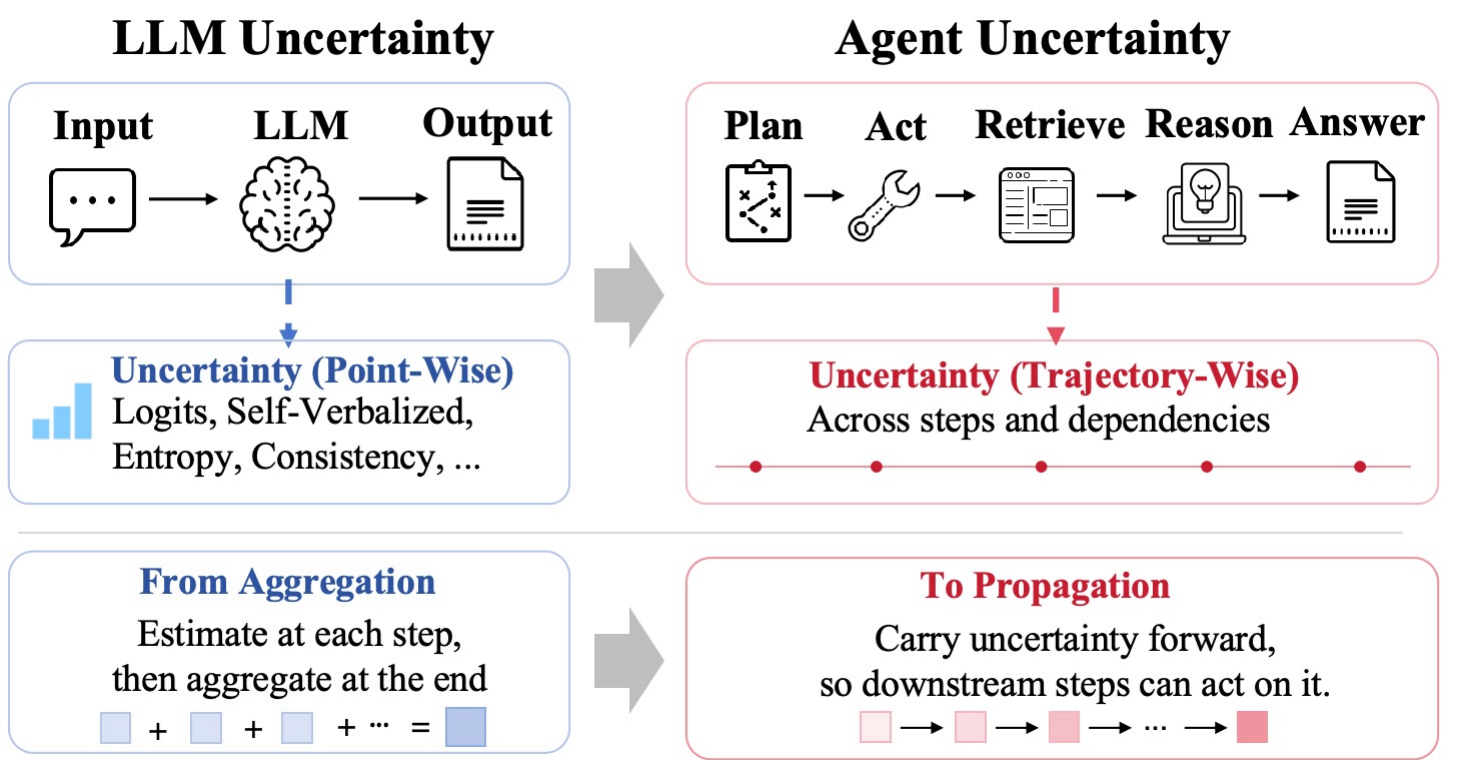}
    \caption{
    Uncertainty Quantification Methods: from LLMs era to Agent era.
    }
    \label{fig:uncertainty-llm-agent}
\end{figure*}

\begin{table*}[tp]
\centering
\small
\renewcommand{\arraystretch}{1.2}
\setlength{\tabcolsep}{8pt}

\begin{tabular}{p{1.5cm} p{1.6cm} p{6.6cm} p{1.5cm}}
\toprule
\textbf{Level} & \textbf{Category} & \textbf{Key Methodology} & \textbf{Works} \\
\midrule

\multirow{10}{*}{\shortstack{\textbf{Atomic}\\ (Single-Step)}} 
& \multirow4{*}{\shortstack{\textbf{Logit-}\\ \textbf{Based}}} 
& \textbf{Logit-based Metrics}: Entropy, MaxProb, etc. for token-level confidence extraction.
& \cite{wu2025search, fadeeva2023lm, kadavath2022language,quevedo2024detecting} \\
\cmidrule(lr){3-4}
& 
& \textbf{Distributional Analysis}: KL-Divergence, Mahalanobis distance, etc.  for latent state deviance.
& \cite{zhao2025learning,xie2025unlocking} \\

\cmidrule(lr){2-4}

& \multirow{2}{*}{\shortstack{\textbf{Self-} \\ \textbf{Verbalized}}} 
& \textbf{Linguistic Reflection}: Direct and multi-pass self-reporting of confidence scores via prompting.
& \cite{xiong2023can,wang2022self,liu2024can} \\

\cmidrule(lr){2-4}

& \multirow{4}{*}{\shortstack{\textbf{Consistency-}\\ \textbf{based}}} 
& \textbf{Semantic Invariance}: Entropy and clustering-based response stability analysis.
& \cite{zheng2025first,dong2025agentic, han2024towards, feng2024rethinking} \\
\cmidrule(lr){3-4}
& 
& \textbf{Ensemble Variance}: Disagreement analysis across multiple candidates or reward model outputs.
& \cite{banerjee2024towards,houliston2024uncertainty,jiang2025vcrl,sun2025uncertainty} \\

\midrule

\multirow{7}{*}{\shortstack{\textbf{Holistic} \\(Trajectory)}} 
& \multirow{2}{*}{\textbf{Aggregation}} 
& \textbf{Step-wise Pooling}: Hierarchical fusion of entropy or variance across the entire execution chain.
& \cite{zhao2024saup, tayebati2026tracer, zhang2026selaur} \\

\cmidrule(lr){2-4}

& \multirow{3}{*}{\textbf{Propagation}} 
& \textbf{Uncertainty Propagation}: Modeling how uncertainty transfers, accumulates, or amplifies across agent steps.
& \cite{zhao2025uncertainty, duan2025uprop} \\

\cmidrule(lr){2-4}

& \multirow{2}{*}{\shortstack{\textbf{Consistency-}\\ \textbf{based}}} 
& \textbf{Path Diversification}: End-to-end consistency analysis across multiple rollout trajectories.
& \cite{stoisser2025towards, wang2024soft, mehta2026agents} \\
\bottomrule
\end{tabular}

\caption{A taxonomy of agent uncertainty estimation methods.}
\label{tab
:refined_taxonomy}
\end{table*}

\section{From Uncertainty Propagation to Interface Bottlenecks}

Uncertainty estimation has moved from isolated language-model predictions to autonomous agent trajectories.
In the LLM setting, uncertainty is usually estimated for a single prediction, such as a next token, generated sentence, retrieved answer, or final response~\citep{shorinwa2025survey,Liu2025UncertaintyQA},
using signals such as entropy~\citep{malinin2021uncertainty,kuhn2023semantic,farquhar2024detecting,nikitin2024kernel,zhang2026semantic},
confidence~\citep{lin2022teaching,tian2023just,xiong2023can},
consistency~\citep{manakul2023selfcheckgpt,wang2022self,lin2023generating},
probing~\citep{azaria2023internal,kossen2024semantic},
or self-evaluation~\citep{kadavath2022language}.
These methods remain useful for estimating local risk at individual agent steps, including tool selection, query formulation, observation interpretation, and intermediate generation~\citep{han2024towards}.

However, they are insufficient for agent systems because agent behavior is trajectory-dependent.
An uncertain early decision can become the premise for later reasoning, tool use, and evidence selection, and small errors at early steps tend to cascade into compounding failures downstream~\citep{zhu2025llm,bao2026drift}.
Thus, agent uncertainty is not only a collection of single-step confidence scores, but a trajectory-level problem in which uncertainty must be modeled across the sequence of decisions that produce the final outcome.

\subsection{Uncertainty is Not Self-Propagating}

The trajectory view clarifies where uncertainty matters, but it does not explain how uncertainty is carried from one component to the next. If a downstream component only receives the output of an upstream decision, then it receives a consequence of uncertainty rather than uncertainty itself. The artifact may have been produced under hesitation, ambiguity, or incomplete evidence, but these conditions are no longer visible once the artifact is exposed as a tool call, query, retrieved passage, rationale, or intermediate answer~\citep{jiang2023active,Ren2023RobotsTA}. The downstream component can continue from the artifact, but it cannot directly access the fragility behind it.

This distinction separates uncertainty propagation from uncertainty re-estimation. Propagation means that some uncertainty-bearing representation survives the handoff and remains usable by the next component, in the sense that uncertainty from earlier decisions is inherited rather than discarded~\citep{duan2025uprop,zhao2024saup}. Re-estimation means that the downstream component sees only a committed artifact and must infer risk after the fact. These are not equivalent. A downstream agent may later notice that a retrieved passage is weak, a tool result is inconsistent, or an intermediate answer conflicts with new evidence, but this does not recover the upstream ambiguity that shaped the original decision. The system is then reacting to traces of uncertainty rather than inheriting uncertainty itself.

This is why trajectory-level estimation alone is insufficient. A trajectory can record where uncertainty appears and how later states become unreliable~\citep{zhao2024saup}, but it does not explain whether uncertainty is preserved when one step becomes input to the next. When interfaces export only discrete commitments, uncertainty may still affect the trajectory causally, but it is no longer available operationally. In that case, the system suffers from upstream uncertainty without receiving it, and propagation collapses into post-hoc reconstruction.

\begin{figure*}[t]
    \centering
    \includegraphics[width=\textwidth]{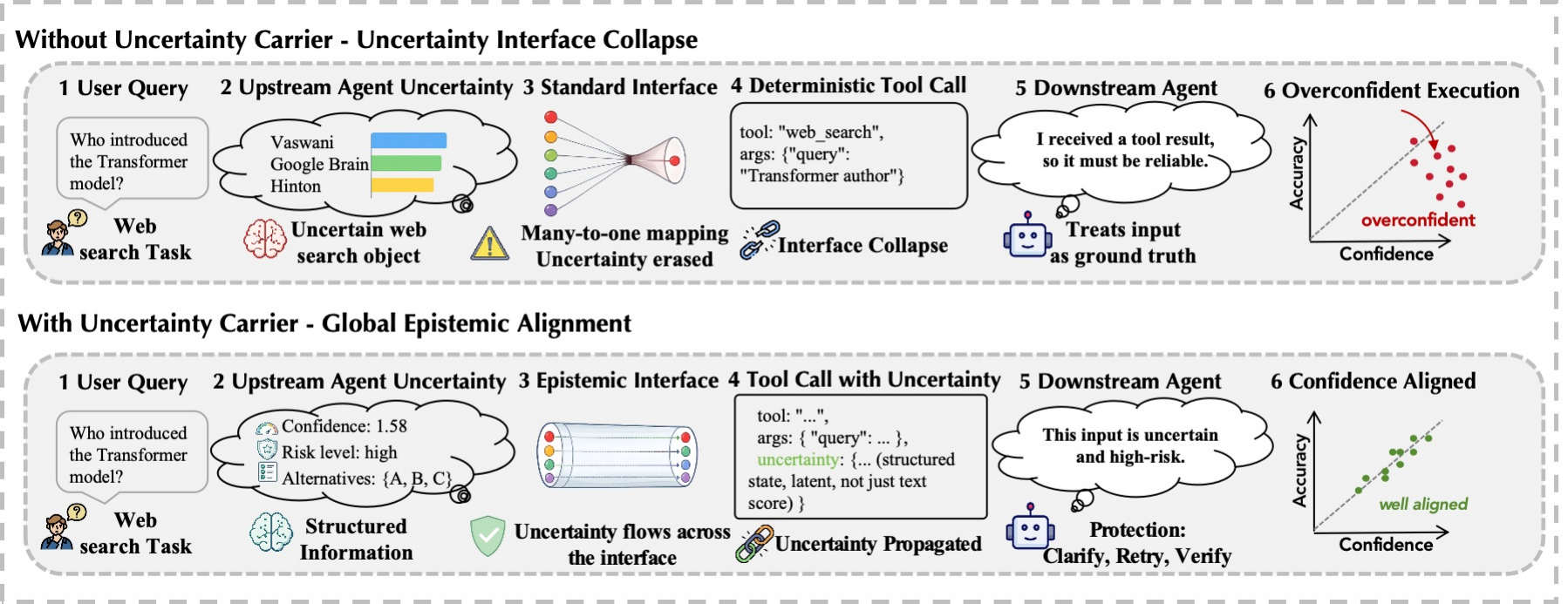}
    \caption{
    Overview of uncertainty interface collapse and uncertainty-preserving handoff.
    Without an uncertainty carrier, upstream uncertainty is erased by the standard interface and downstream agents may execute with unwarranted confidence.
    With an uncertainty carrier, uncertainty flows across the interface, allowing downstream agents to recognize high-risk inputs and trigger recovery actions such as clarification, retry, and verification.
    }
    \label{fig:interface-collapse-overview}
\end{figure*}

\subsection{Interfaces Expose Commitments, Not Epistemic States}

The reason uncertainty fails to propagate is that standard agent interfaces do not transmit epistemic states. They transmit commitments. A tool interface requires a tool name and arguments~\citep{schick2023toolformer}; a retrieval interface requires a query~\citep{jiang2023active}; a structured-output interface requires fields conforming to a fixed schema; an action interface requires a selected next step~\citep{yao2023react}. These formats make modular agent systems executable, but they also force an uncertainty-bearing internal state to be projected into a single consumable artifact.

\paragraph{The Trap: A many-to-one projection}
 Before acting, an upstream agent may distribute belief over several possible actions, queries, evidence sources, or interpretations. The interface does not expose this distribution. It records only the selected action. As a result, different upstream states with very different uncertainty profiles can produce the same tool call, query string, retrieved passage, or intermediate answer. A near-tie among several actions and a confident decision may therefore become indistinguishable once they cross the interface.

\begin{claimbox}
\textbf{Interface Collapse.}
Standard agent interfaces perform a many-to-one projection from uncertain internal states to discrete commitments. Different epistemic states can therefore produce the same artifact, causing downstream components to receive the selected decision without the uncertainty profile behind it.
\end{claimbox}

This is the point at which interface collapse occurs. The downstream component receives the artifact, but not the instability behind it. It can observe what the upstream component committed to, but not the competing alternatives, weak evidence, unresolved ambiguity, or confidence gap that shaped the commitment. The interface therefore preserves the decision while discarding the epistemic context under which it was made. In this sense, uncertainty is not merely reduced; it is structurally erased by the commitment format itself.

\subsection{The Interface Bottleneck}

\begin{wrapfigure}{r}{0.4\linewidth}
\vspace{-8pt}
\centering
\includegraphics[width=\linewidth]{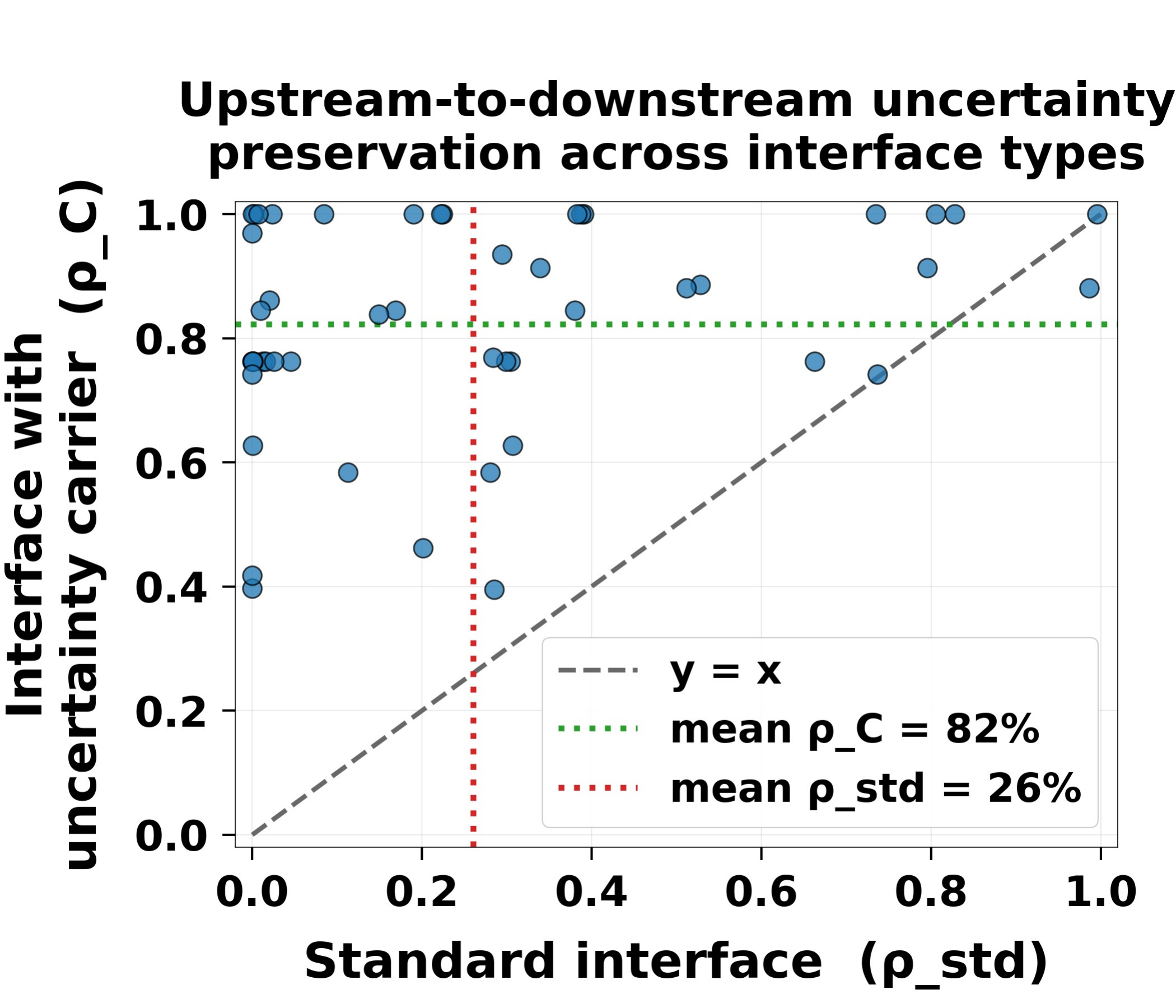}
\caption{
Interface uncertainty loss without an uncertainty carrier.
}
\label{fig:interface-bottleneck}
\vspace{-8pt}
\end{wrapfigure}

The many-to-one projection described above creates an interface bottleneck in agent uncertainty propagation. Once an uncertainty-bearing state is compressed into a committed artifact, uncertainty may still influence the trajectory, but it no longer travels as an accessible state. It survives only indirectly, through weak evidence, brittle plans, inconsistent tool results, unsupported intermediate answers, or later downstream failures. This bottleneck separates uncertainty propagation from post-hoc risk detection. A downstream component may estimate uncertainty over the artifact it receives, but this is not equivalent to receiving the upstream uncertainty that produced that artifact. After interface collapse, downstream reasoning begins from an incomplete premise: the system sees the decision, but not the fragility behind the decision. It may still detect risk later, but it detects risk after the handoff rather than preserving uncertainty through the handoff.

The central challenge is therefore not only to measure uncertainty at each step, but to decide what form of uncertainty should cross the interface. Without an uncertainty-bearing carrier, agent systems pass decisions forward while leaving doubt behind.

\begin{claimbox}
\textbf{Interface Bottleneck.}
Once an uncertainty-bearing state is compressed into a committed artifact, uncertainty may still shape the trajectory causally, but it no longer travels as an accessible state for downstream recovery.
\end{claimbox}

\section{Uncertain Decision Handoff and Confidence Laundering}
The interface bottleneck explains how uncertainty is lost at the point of handoff. We now characterize the failure mode that follows from this loss: unresolved uncertainty is converted into an executable commitment, and that commitment is later treated as a stable premise.

\subsection{Premature Commitment}
An upstream agent may maintain multiple plausible hypotheses, conflicting evidence, or unstable action candidates~\citep{wang2022self, yao2023tot}, but the interface requires one action, one query, one answer, or one structured field~\citep{yao2023react,pham2024cipher}. This commitment is necessary for execution because downstream components need concrete objects to consume. Yet it is risky because the artifact enters the downstream context before the uncertainty behind it has been resolved.

\subsection{Uncertain Decision Handoff}

We define an \textit{uncertain decision handoff} as the process by which an upstream component commits to an intermediate decision under uncertainty and transfers the resulting artifact to a downstream component for continued reasoning or action. This is not ordinary communication. Communication shares information; handoff transfers operational responsibility. The downstream component may execute the artifact, verify it, revise it, or use it as a premise for the next step.

This framing matters because the artifact and the uncertainty behind it are separable. A handoff may preserve the appearance of progress while removing the epistemic context needed to decide whether to trust, verify, retry, clarify, or reroute.

\begin{claimbox}
\textbf{Uncertain Decision Handoff.}
A handoff is uncertainty-sensitive because it transfers not only information, but operational responsibility. If the artifact crosses the interface without its decision fragility, downstream components inherit a usable object without the context needed to judge how it should be used.
\end{claimbox}

Common handoffs show this pattern directly: a tool call can hide uncertainty about tool suitability, a query can hide uncertainty about the intended information need, a retrieved passage can hide uncertainty about evidence coverage, a selected plan can hide uncertainty about path reliability, and a structured field can hide uncertainty about whether the extracted value is correct.

\subsection{Confidence Laundering}

\begin{figure*}[t]
    \centering
    \includegraphics[width=0.95\textwidth]{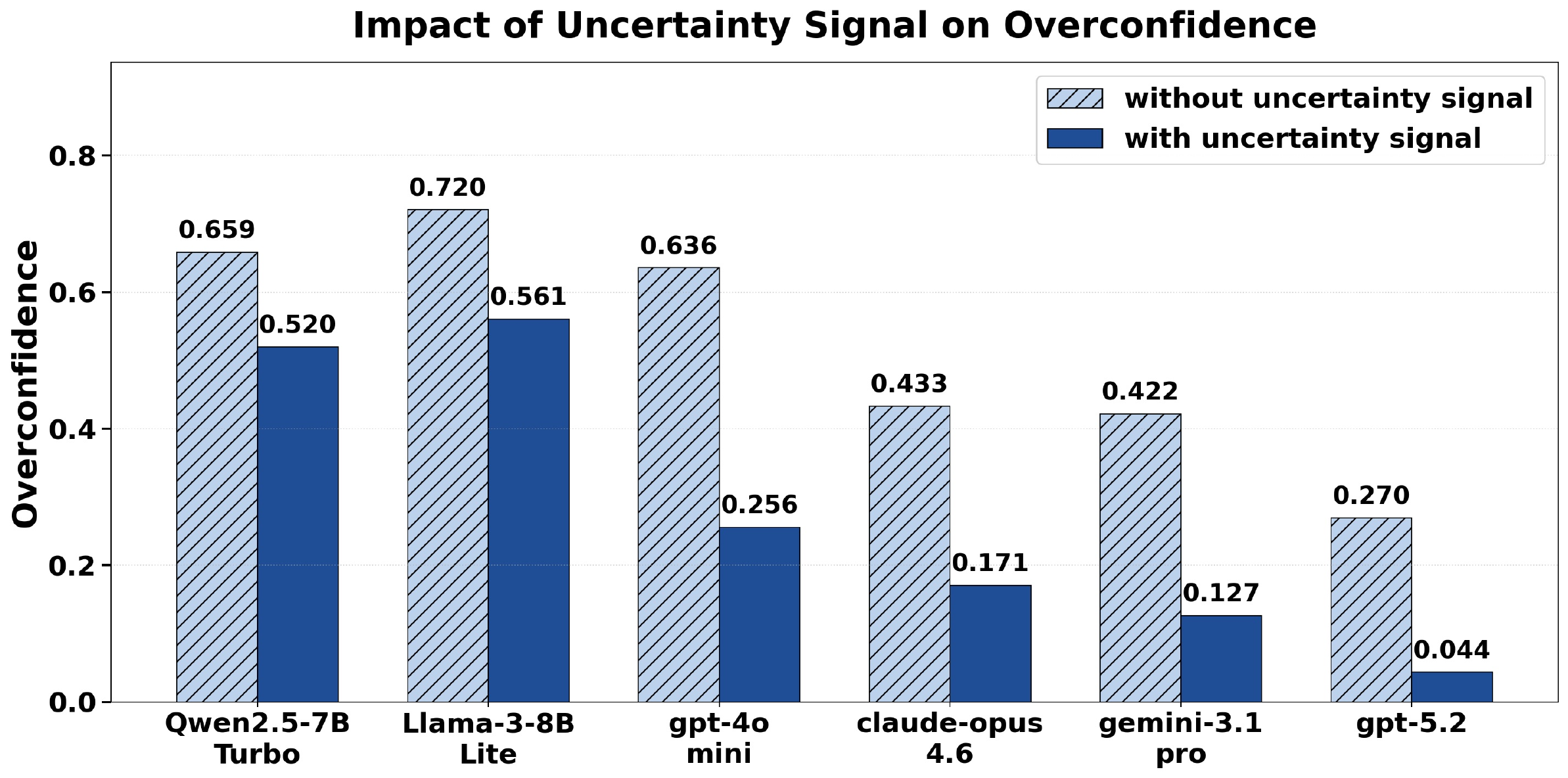}
    \caption{
    Impact of uncertainty signals on overconfidence across different language models.
    For each model, we compare downstream overconfidence with and without an uncertainty signal.
    Across all evaluated models, uncertainty information consistently reduces overconfidence, suggesting that downstream components are less likely to over-trust upstream outputs when uncertainty is preserved across the handoff.
    This result supports our claim that uncertainty carriers can mitigate interface-induced overconfidence in agent systems.
    }
    \label{fig:overconfidence}
\end{figure*}

We call the downstream effect \textit{confidence laundering}: a fragile upstream decision is packaged as a clean symbolic or executable artifact and becomes easier to over-trust. The interface does not merely remove uncertainty; it gives the decision a form that appears procedurally valid. A complete API call appears executable, a fluent intermediate answer appears reasoned, a structured JSON field appears validated, a returned passage appears relevant, and a verification judgment appears authoritative.

\begin{claimbox}
\textbf{Confidence Laundering.}
Interface collapse can transform unresolved upstream uncertainty into apparent downstream confidence by presenting fragile decisions as polished, executable, or formally valid artifacts.
\end{claimbox}

The danger is that procedural validity can be mistaken for epistemic reliability. Once this happens, errors are no longer local. A fragile upstream decision becomes a downstream premise, that premise supports further reasoning, and the system may proceed with increasing confidence along a path that should have been questioned at the handoff.

\section{Latent Uncertainty as a Carrier}

The previous section identified the failure mode: agent interfaces can turn fragile upstream decisions into clean artifacts that downstream components over-trust. This suggests that the remedy is not simply to estimate uncertainty more accurately after the fact. What is missing is a carrier: a representation that travels with the artifact and preserves the epistemic fragility behind it. We use \textit{latent uncertainty} to name this object. The goal is not to replace textual communication with hidden states, nor to claim that latent representations are always superior. The goal is narrower: to preserve upstream decision fragility in a downstream-usable form before interface collapse becomes cascading failure.

\subsection{Definition of Latent Uncertainty}

\paragraph{Uncertainty-bearing internal representations.}
We define \textit{latent uncertainty} as uncertainty-bearing information encoded in an agent's internal continuous representations, such as hidden states, activations, or KV traces, that is not fully preserved by the committed text or action interface but can be attached to a handoff in a downstream-usable form. Before commitment, an agent's internal state may contain competing hypotheses, hidden disagreement, prompt or context sensitivity, unstable tool choice, evidence conflict, ambiguous intent, or uncertainty about the source of risk. After commitment, these signals may no longer appear in the emitted tool call, query, retrieved evidence, structured field, or intermediate answer.

\paragraph{Actionable subset of hidden states.}
Latent uncertainty is therefore narrower than ``all information in a hidden state.'' Not every hidden-state feature is uncertainty-relevant, and not every latent signal supports downstream control. We use the term to refer only to the subset of internal representation that helps a downstream component make uncertainty-sensitive decisions beyond what is recoverable from the committed artifact alone. It matters only if it improves the downstream component's ability to decide whether to trust, verify, retry, clarify, abstain, or reroute.

\begin{figure*}[t]
    \centering
    \includegraphics[width=\textwidth]{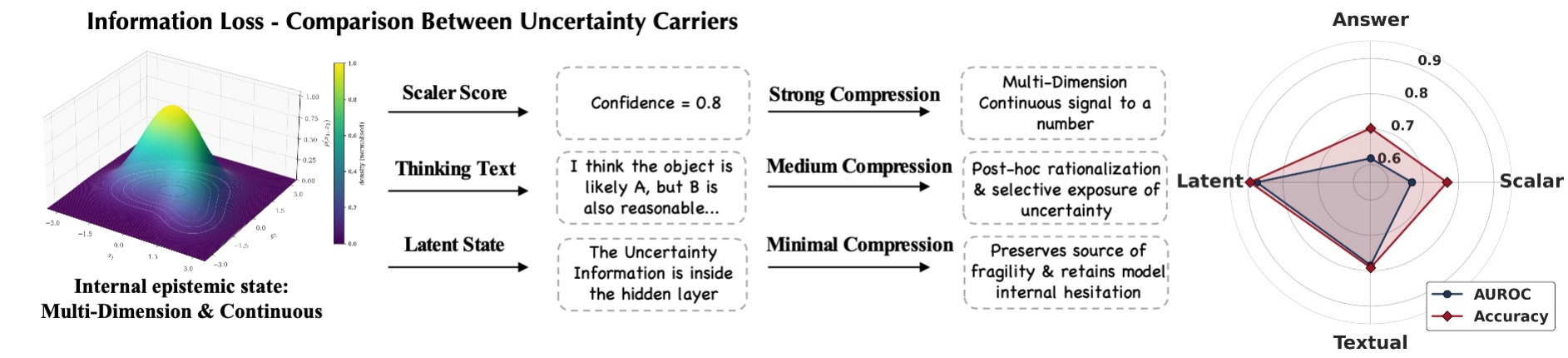}
    \caption{
    Comparison of information loss across different uncertainty carriers.
    Scalar confidence compresses a multi-dimensional epistemic state into a single number, while thinking text exposes uncertainty through a selective natural-language rationale.
    In contrast, latent uncertainty preserves more of the model's internal hesitation and source of fragility, which leads to stronger downstream uncertainty recovery.
    }
    \label{fig:uncertainty-carrier-comparison}
\end{figure*}

\subsection{Why Scalar Confidence is Insufficient}

The carrier view clarifies why scalar confidence is only a partial solution. A confidence score annotates a committed artifact, but it does not necessarily preserve the epistemic state that produced it. It can indicate that an output is risky, but often cannot identify why it is risky, which recovery action is appropriate, or how the risk should interact with later decisions.

This limitation is structural. A single number collapses different sources of uncertainty into one degree of trust. It is also hard to compose across multi-step systems, where risk may arise from tool choice, retrieval coverage, ambiguous intent, evidence conflict, or reasoning instability. Verbal explanations add readability, but they are phrasing-sensitive and not necessarily calibrated for downstream use. Sample summaries expose output variation, but they can be expensive and still require the downstream component to infer the right recovery action.

Latent uncertainty instead aims to travel with the artifact as a representation-level trace of the epistemic state that produced it, while still accepting its costs in interpretability and interface simplicity.

\subsection{Attaching Latent Uncertainty to Handoffs}

Once latent uncertainty is understood as a carrier, the design question is how much of the upstream epistemic state should cross the handoff. A lightweight option is to probe hidden states into calibrated risk scores, which is easy to use but collapses the latent signal back into a scalar. A more expressive option is to encode upstream uncertainty into a small number of soft-prompt vectors appended to the downstream model's input, allowing the downstream component to condition on upstream fragility. The strongest form is KV or activation handoff, where part of the upstream uncertainty-bearing state is passed directly to the downstream component. These designs differ in cost, interpretability, and compatibility, but they share the same principle: the handoff should contain not only what was decided, but how fragile that decision was.

\subsection{Difference from Latent Communication}
This principle separates latent uncertainty from existing latent communication. Both use internal representations as a communication medium, but they answer different questions. Existing latent communication treats latent states as substitutes for natural language messages, motivated by bandwidth, efficiency, or task performance~\citep{pham2024cipher,hao2024coconut,zou2025interlat}. Hidden states, activations, or KV traces may let agents exchange richer information with fewer tokens~\citep{shi2025kvcomm}.

Our motivation is epistemic preservation. We are not asking whether agents can communicate more information without text. We ask whether uncertainty can survive the moment when an upstream agent commits to an action, query, retrieved evidence, structured output, or intermediate answer. This changes the purpose of the channel from transmitting more content to preserving the epistemic status of a decision. Latent communication asks how agents can share more information~\citep{pham2024cipher,zou2025interlat}. Latent uncertainty asks how agents can avoid laundering uncertainty into confidence. The former is primarily a communication-efficiency problem; the latter is an interface-reliability problem. We therefore shift latent communication from sharing thoughts to preserving doubt.

\section{Empirical Observation}
\label{sec:empirical-validation}

\begin{figure*}[t]
    \centering
    \includegraphics[width=\linewidth]{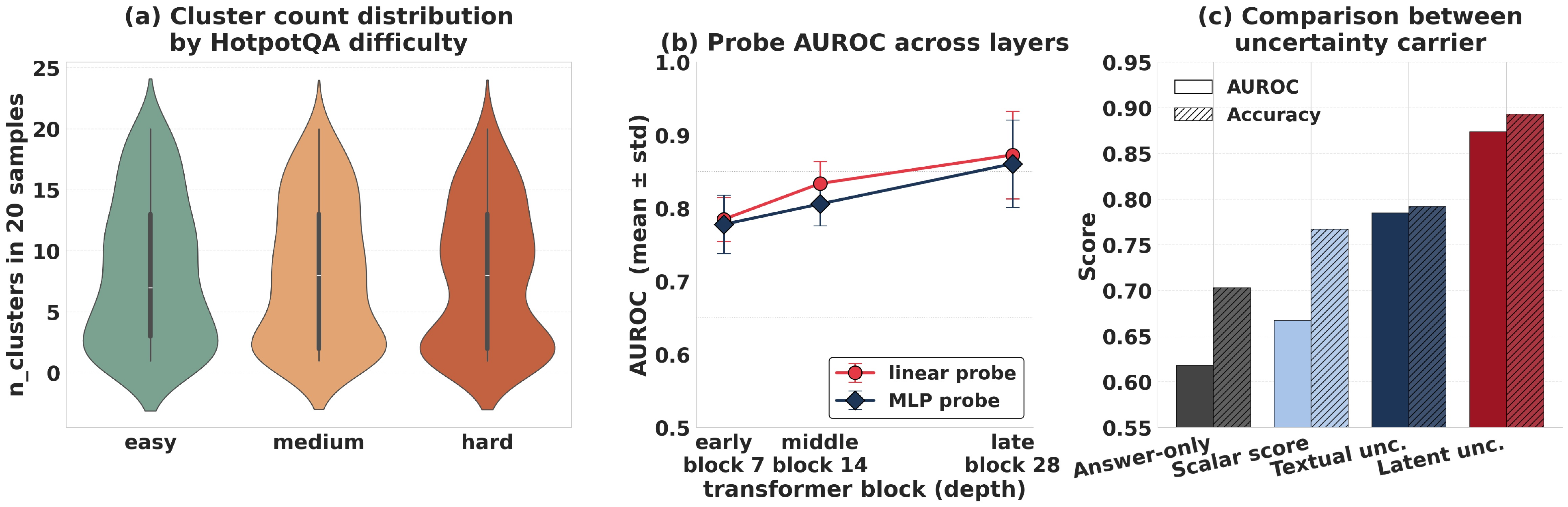}
    \caption{
    Empirical evidence for latent uncertainty as an uncertainty carrier.
    (a) Semantic cluster-count distributions overlap across HotpotQA difficulty levels, showing that external difficulty labels do not fully explain Qwen-1.5B's uncertainty behavior.
    (b) Hidden-state probes recover semantic-entropy-based uncertainty labels with strong AUROC, and performance improves across transformer depth.
    (c) Latent uncertainty outperforms answer-only, scalar-score, and textual-uncertainty interfaces, suggesting that representation-level carriers better preserve upstream fragility for downstream use.
    }
    \label{fig:combined_results}
\end{figure*}

Our position makes two targeted empirical claims: model uncertainty cannot be reliably inferred from external task labels alone, and internal latent states preserve uncertainty-relevant information that is partly lost when the interface exposes only a committed artifact. To test these claims, we use Qwen-1.5B on HotpotQA. For each question, we sample multiple answers, cluster them semantically, and derive uncertainty labels from the resulting cluster structure. We then compare how well uncertainty can be recovered from different handoff signals: answer-only, scalar score, textual uncertainty, and latent uncertainty derived from upstream hidden representations.

Figure~\ref{fig:combined_results} provides three observations. \textbf{First, HotpotQA difficulty labels do not cleanly predict model uncertainty:} easy, medium, and hard questions show substantially overlapping semantic-cluster distributions. This suggests that the downstream component cannot recover uncertainty simply by relying on external task difficulty; the uncertainty signal is model-specific rather than reducible to dataset-level difficulty labels. 

\textbf{Second, hidden states contain recoverable uncertainty information:} lightweight probes predict semantic-entropy-based labels with strong AUROC, and performance improves across depth. This indicates that uncertainty is encoded in the model's latent computation before it is compressed into a final textual answer. 

\textbf{Third, latent uncertainty provides the strongest downstream signal,} outperforming answer-only, scalar-score, and textual-uncertainty interfaces in AUROC and accuracy. This supports the interface-bottleneck view: committed textual artifacts lose upstream fragility, while hidden representations preserve uncertainty signals that can be passed downstream.

\section{Discussion}

Our position reframes agent uncertainty propagation as an interface problem. If uncertainty is lost during handoff, better step-wise estimation is not sufficient: downstream components cannot act on uncertainty that never reaches them. Agent systems therefore need uncertainty-preserving interfaces, not only post-hoc uncertainty estimates.

\subsection{Limitations}

\paragraph{Scope of the claim.}
Our argument is not that latent uncertainty should replace existing uncertainty signals. Scalar confidence, textual rationales, and explicit uncertainty schemas remain useful when risk is local, simple, or easy to verbalize. Latent uncertainty is most relevant when downstream behavior depends on fragility that is difficult to recover from the committed artifact alone.

\paragraph{Practical constraints.}
Latent carriers may require access to model internals, additional interface design, and compatibility between upstream and downstream components. These requirements may limit their use in black-box or heterogeneous agent systems.

\paragraph{Interpretability.}
Latent uncertainty may preserve richer information, but it is less interpretable than scalar or textual signals. In practice, latent carriers should be paired with interpretable summaries or recovery policies when human inspection is needed.

\subsection{Alternative Views}

\paragraph{Downstream re-estimation.}
One alternative is to let downstream components re-estimate uncertainty from the artifact they receive. This is useful and often necessary, but it operates after the handoff. It may detect that an answer, query, or tool result is risky, but it cannot fully recover the upstream epistemic state that produced that artifact.

\paragraph{Explicit uncertainty annotations.}
Another alternative is to attach confidence scores or natural-language rationales to intermediate artifacts. We view these as useful uncertainty carriers, especially when the relevant risk is simple or easily verbalized. However, they can still compress away the structure of uncertainty, including which part of the upstream decision was fragile and what recovery action is appropriate.

\paragraph{Carrier choice.}
Our claim is not that one carrier is universally optimal. Different agent settings may require different uncertainty interfaces. The practical goal is to preserve enough uncertainty for the next component to decide whether to continue, verify, retry, clarify, abstain, or reroute. Latent uncertainty is one candidate carrier for settings where scalar or textual signals are too lossy.

\subsection{Broader Implications}

\paragraph{Uncertainty should become part of the interface contract.}
Current agent interfaces are optimized for executability: they pass tool names, arguments, queries, passages, plans, fields, or judgments. This makes coordination possible, but it also allows clean artifacts to hide the uncertainty that shaped them. Our view suggests that uncertainty should not be optional metadata added after the fact. It should be part of the contract between components, so that a downstream agent can know whether an artifact was produced confidently, under ambiguity, from weak evidence, or after unstable action selection. In long-horizon systems, a handoff that omits this context may be incomplete even when the artifact is syntactically valid.

\paragraph{Evaluation should move from final accuracy to handoff reliability.}
If uncertainty propagation is an interface problem, evaluation should not stop at final-answer accuracy, calibration, or abstention. Agent benchmarks should test whether downstream components avoid over-trusting fragile upstream decisions and choose appropriate recovery behavior. Metrics such as recovery recall, false accept rate, retry success, clarification precision, confidence-laundering reduction, and error amplification depth ask whether uncertainty actually changes downstream behavior, not merely whether the system can assign a confidence score.

\paragraph{Uncertainty should be linked to recovery actions.}
Agent uncertainty is not a single scalar quantity. Ambiguous intent, weak retrieval coverage, tool mismatch, execution instability, evidence conflict, and reasoning uncertainty require different interventions. Clarification, retrieval, rerouting, verification, and decomposition should be triggered by different uncertainty sources. An uncertainty-preserving interface should therefore communicate not only the degree of risk, but also the kind of risk and the recovery action it supports.

\paragraph{Agent architectures should shift from clean artifacts to recoverable systems.}
The standard modular pipeline treats components as producers and consumers of artifacts: a planner emits a plan, a retriever emits passages, a tool emits results, and a verifier emits a judgment. Our framing suggests a different design goal. Components should exchange artifacts together with enough uncertainty-bearing context for the next component to decide whether to continue, verify, retry, clarify, abstain, or reroute. The objective is not to make every intermediate artifact appear complete, but to preserve doubt long enough for the system to recover from it.

\newpage
\bibliographystyle{unsrtnat}
\bibliography{reference}






\newpage
\appendix

\section{Experimental Details}
\label{app:experimental-details}

\subsection{Task, Tool-Use Setting, and Model}

We use HotpotQA as the evaluation setting because it contains multi-hop questions that often require gathering and combining information from multiple sources. This type of multi-hop question answering is closely related to recent multi-agent QA settings, where different agents or modules coordinate over structured information and intermediate reasoning steps~\citep{shi2026ng}. However, we do not use HotpotQA as a standard retrieval benchmark in which the model is given the original supporting documents or a fixed retrieved context. Instead, we convert HotpotQA into a web-search tool-use setting. Each question is treated as a user query given to an agent that must decide how to search the web, consume external search results, and produce a final answer.

Concretely, the upstream agent is given only the HotpotQA question. It is not given the gold supporting paragraphs, the original Wikipedia context, or an oracle retrieval result. The agent must first formulate a search query and call an external web-search tool. We use the Tavily Web Search API as the tool interface. Tavily returns web search results, including titles, snippets, and source URLs. These returned results serve as the external observation available to the agent. This open-web evidence collection setting is related to recent multi-agent systems for web data collection, where agents must gather, filter, and organize external information before downstream use~\citep{ma2026autodata}. The agent then generates an answer conditioned on the original question and the Tavily search output.

This construction is important for our purpose because it makes the experiment an instance of uncertainty-sensitive tool use rather than static retrieval. The uncertainty of interest does not only come from the final answer. It can also arise when the agent decides how to formulate the search query, whether the search results cover the necessary evidence, whether snippets from different sources are consistent, and whether the available web evidence is sufficient for answering the question. Thus, the setting contains an explicit interface between an upstream decision and an external tool call. This allows us to study whether uncertainty behind the tool-mediated answer is preserved or lost after the agent commits to a final artifact.

Each question is associated with a HotpotQA difficulty label, namely easy, medium, or hard. These labels are used only for analysis in Figure~\ref{fig:combined_results}(a), not as supervision for the uncertainty classifier. Our uncertainty labels are instead derived from the upstream model's own sampled behavior under the web-search tool-use setting.

The upstream model is Qwen-1.5B. We intentionally use a relatively small model because its uncertainty behavior is visible across a broad range of examples. A stronger model may collapse many questions into low-uncertainty behavior, while a much weaker model may behave uncertainly on most questions. Qwen-1.5B provides a useful middle regime in which both low- and high-uncertainty examples are present, allowing us to test whether uncertainty can be recovered from latent representations rather than inferred from task difficulty alone.

\subsection{Web-Search Tool-Use Construction}

We convert HotpotQA into a web-search tool-use setting rather than using it as a static retrieval benchmark. In the original HotpotQA setting, each question is associated with supporting paragraphs from Wikipedia. In our setting, these supporting paragraphs are not exposed to the model. The model receives only the question and must interact with an external web-search tool to obtain evidence before producing an answer.

We use the Tavily Web Search API as the external tool. Given a question, the upstream agent first formulates a natural-language search query. This query-formulation step is related to broader work on prompt and agent optimization, where the behavior of an agent depends strongly on how intermediate prompts or instructions are selected and refined~\citep{zhang2026mapro}. Tavily then returns web-search results, including titles, snippets, and source URLs. These returned results are treated as the tool observation. The agent produces its final answer conditioned on the original question and the Tavily observation.

This construction turns HotpotQA from a retrieve-and-read task into a tool-mediated decision process. The uncertainty of interest may arise from several stages: the search query may be underspecified, the returned web evidence may be incomplete, different snippets may point to competing answers, or the model may be uncertain about whether the available evidence is sufficient. Such decisions are also connected to agent-routing settings, where systems must choose or adapt the path through which a question is handled before reaching a final answer~\citep{huang2026evolverouter}. These sources of fragility are precisely what standard committed interfaces tend to hide.

\begin{bluecasebox}{Case A.2: Converting HotpotQA into Web-Search Tool Use}
\textbf{Original HotpotQA format.}  
The model is evaluated on a multi-hop question, often with gold supporting facts or a fixed corpus available for retrieval.

\medskip
\textbf{Our tool-use format.}  
The model receives only the question. It is not given the gold supporting paragraphs or oracle evidence. It must first produce a search query, call Tavily, read the returned web snippets, and then produce a final answer.

\medskip
\textbf{Why this matters.}  
The experiment is no longer only about whether the model can answer from a retrieved passage. It also tests whether uncertainty from the tool-use process survives the handoff. The agent may be uncertain about the query, the evidence coverage, or the reliability of the returned snippets, but the downstream component may receive only the final answer.
\end{bluecasebox}

\subsection{Prompt Format}

We use a fixed prompt template for the web-search agent. The prompt asks the model to identify a search query, use the returned Tavily evidence, and produce a concise answer. The purpose is not to optimize web-search performance, but to create a controlled interface where a question is transformed into a tool call and then into a committed answer.
\subsection{Illustrative Example}

The following example illustrates how a HotpotQA question is converted into a tool-use trajectory. The exact question and returned search results are illustrative; the purpose is to show the structure of the interface rather than to introduce a new benchmark example.

\subsection{Sampling and Semantic Clustering}

For each question, we sample multiple tool-use trajectories from the upstream model using the same prompt format. Let $K$ denote the number of sampled outputs. In our main experiment, we use $K=20$ samples per question. Each sample corresponds to a complete web-search-based answer generation process: the model receives the question, uses the Tavily search result as the external observation, and produces a final answer.

\begin{yellowcasebox}{Case A.3: Prompt Template for the Web-Search Agent}
\textbf{System instruction.}

You are a question-answering agent with access to a web-search tool.  
Given a question, first write a search query that would help answer the question.  
After receiving web-search results, answer the question using only the provided search evidence.  
If the evidence is incomplete or conflicting, still provide the best answer but do not invent unsupported facts.

\medskip
\textbf{User input.}

Question: \texttt{\{hotpotqa\_question\}}

\medskip
\textbf{Tool call format.}

Search query: \texttt{\{model\_generated\_query\}}

\medskip
\textbf{Tool observation.}

Tavily returns a list of search results:
\begin{itemize}
    \item title
    \item snippet
    \item source URL
\end{itemize}

\medskip
\textbf{Final answer format.}

Answer: \texttt{\{final\_answer\}}
\end{yellowcasebox}

\begin{greencasebox}{Case A.4: Example Tool-Use Trajectory}
\textbf{HotpotQA question.}

Which film was released earlier, the film directed by the director of \textit{Inception}, or the film starring the actor who played Jack Sparrow?

\medskip
\textbf{Step 1: Search-query decision.}

The agent must decide what to search for. Several queries are possible, such as:
\begin{itemize}
    \item \texttt{director of Inception filmography release dates}
    \item \texttt{actor who played Jack Sparrow films release dates}
    \item \texttt{Inception director and Jack Sparrow actor comparison}
\end{itemize}

The selected query is a committed artifact. It hides whether the agent was confident in this query or whether several competing query formulations were plausible.

\medskip
\textbf{Step 2: Tavily web-search call.}

The agent calls Tavily with the selected query. Tavily returns web snippets and source URLs. These snippets may identify Christopher Nolan as the director of \textit{Inception} and Johnny Depp as the actor who played Jack Sparrow.

\medskip
\textbf{Step 3: Evidence use.}

The agent reads the returned snippets and infers the relevant films and release years. If the snippets are incomplete, the agent may still produce an answer based on partial evidence.

\medskip
\textbf{Step 4: Final committed answer.}

The agent emits a concise answer. Downstream components receive this answer as a clean artifact, but they may not receive the uncertainty behind the query choice, evidence sufficiency, or source consistency.
\end{greencasebox}

Each sampled final answer is normalized before clustering. Normalization removes superficial variation such as capitalization, punctuation, and minor formatting differences, while preserving semantic answer content. We then group the sampled answers into semantic clusters. Each cluster corresponds to a distinct answer meaning rather than an exact surface form. The number and distribution of these clusters provide a behavioral measure of model uncertainty under tool use.

If most sampled answers fall into a single semantic cluster, the model exhibits stable answer behavior under the web-search setting. If the samples spread across several clusters, the model exhibits unstable or uncertain behavior. This instability may reflect answer ambiguity, weak search evidence, sensitivity to search results, or uncertainty in how the model uses the web-search observation. We therefore treat semantic clustering as a behavioral proxy for the uncertainty of the complete tool-mediated decision process, rather than only the uncertainty of a closed-book answer.

\subsection{Uncertainty Labels from Semantic Entropy}

We derive uncertainty labels from the semantic cluster distribution. Let the sampled outputs for a question form semantic clusters $\mathcal{C}=\{c_1,\ldots,c_m\}$, where each cluster $c_i$ contains $n_i$ samples and $\sum_i n_i=K$. We define the empirical cluster probability as
\[
p_i = \frac{n_i}{K}.
\]
The semantic entropy of the model's sampled behavior is then
\[
H_{\mathrm{sem}} = - \sum_{i=1}^{m} p_i \log p_i.
\]
Questions with low semantic entropy are treated as low-uncertainty examples, while questions with high semantic entropy are treated as high-uncertainty examples. In the binary setting used for probe evaluation, we threshold semantic entropy to construct uncertainty labels. This label is not based on the dataset difficulty annotation. It is derived from the model's own output distribution in the Tavily-based web-search setting, making it a behavioral measure of the upstream agent's uncertainty.

\subsection{Hidden-State Extraction}

To test whether uncertainty is encoded in latent representations, we extract hidden states from selected transformer blocks of the upstream model during answer generation. We consider early, middle, and late layers, corresponding to transformer blocks 7, 14, and 28 in Figure~\ref{fig:combined_results}(b). For each question, we collect the hidden representation associated with the model's final answer behavior after it has consumed the web-search observation. These representations serve as candidate latent uncertainty carriers.

The purpose of this analysis is not to claim that a particular layer is universally optimal. Rather, the goal is to test whether uncertainty-relevant information is recoverable from internal states and whether this recoverability changes with model depth. In our setting, such information may reflect not only answer uncertainty, but also uncertainty induced by the tool-use process, including search-result coverage, evidence conflict, and sensitivity to the external observation.

\subsection{Probe Training}

We train lightweight probes to predict the semantic-entropy-based uncertainty label from hidden representations. We evaluate two probe families. The first is a linear probe implemented as logistic regression. The second is an MLP probe with nonlinear hidden layers. Both probes take the extracted hidden representation as input and output a binary uncertainty prediction.

The probes are trained on a training split and evaluated on held-out examples. We report AUROC as the main metric because it measures ranking quality across thresholds and is appropriate for uncertainty detection. We also report standard deviation across repeated splits or seeds when available. The results in Figure~\ref{fig:combined_results}(b) show that both probe types recover uncertainty information from hidden states, with stronger performance in later layers.

\subsection{Handoff Conditions}

To evaluate latent uncertainty as a carrier, we compare several downstream handoff conditions. Each condition represents a different amount or form of information passed from the upstream tool-use agent to the downstream uncertainty predictor.

\paragraph{Answer-only.}
The downstream component receives only the committed final answer. This condition represents a standard handoff in which the upstream agent exposes the answer produced after web search, but not the uncertainty state that produced it. It therefore hides uncertainty about search-query formulation, evidence coverage, and answer stability.

\paragraph{Scalar score.}
The downstream component receives the committed answer together with a scalar uncertainty score. This condition represents uncertainty annotation. It tests whether a compact confidence-like signal is sufficient to recover upstream fragility after the web-search tool-use process.

\paragraph{Textual uncertainty.}
The downstream component receives the committed answer together with a textual uncertainty description. This condition represents a more expressive but still text-based uncertainty interface. It can expose selected reasons for doubt, but it remains a compressed natural-language account of the upstream uncertainty state.

\paragraph{Latent uncertainty.}
The downstream component receives the committed answer together with a latent uncertainty carrier derived from the upstream hidden representation. This condition tests whether representation-level uncertainty preserves information that is not available from the answer alone or from coarse textual annotations. In the web-search setting, this latent signal may encode fragility related to ambiguous search intent, weak or conflicting snippets, unstable evidence use, or uncertain answer generation.
\subsection{Uncertainty Lost at the Tool-Use Handoff}

In this setting, the key interface is the transition from an uncertainty-bearing internal state to a committed tool-use artifact. The agent may internally represent uncertainty about what to search, which evidence to trust, and whether the returned results are sufficient. However, the standard interface exposes only the selected search query, the returned snippets, and the final answer. This creates a natural site for interface collapse.

\begin{graycasebox}{Case A.10: What the Standard Interface Hides}
\textbf{Before the handoff.}  
The upstream agent may be uncertain about several aspects of the task:
\begin{itemize}
    \item whether the question requires one search or multiple searches;
    \item whether the generated query captures the intended entity relation;
    \item whether Tavily returned enough evidence to support the answer;
    \item whether different snippets imply conflicting answers;
    \item whether the final answer is sensitive to one weak source.
\end{itemize}

\medskip
\textbf{After the handoff.}  
The downstream component usually receives only:
\begin{itemize}
    \item the selected search query;
    \item the Tavily search results;
    \item the final answer;
    \item optionally, a scalar confidence score or textual uncertainty statement.
\end{itemize}

\medskip
\textbf{Interface-bottleneck interpretation.}  
The downstream component can inspect the committed artifacts, but it cannot directly recover the upstream uncertainty that shaped them. This is why we compare answer-only, scalar-score, textual-uncertainty, and latent-uncertainty handoffs.
\end{graycasebox}
\subsection{Evaluation Metrics}

We report AUROC and accuracy for the comparison between uncertainty carriers. AUROC measures how well each handoff condition ranks high-uncertainty examples above low-uncertainty examples. Accuracy measures binary classification performance after thresholding. These metrics correspond to the downstream component's ability to detect whether the upstream tool-mediated decision is fragile.

In the context of our position, these metrics should be interpreted as measures of uncertainty recoverability after the handoff. A low score indicates that the handoff signal does not preserve enough upstream uncertainty for downstream use. A high score indicates that the signal functions as a more effective uncertainty carrier.

\subsection{Interpretation}

The results support three observations. First, HotpotQA difficulty labels are not sufficient to explain Qwen-1.5B's uncertainty behavior in the web-search tool-use setting, since semantic cluster-count distributions overlap substantially across easy, medium, and hard examples. Second, hidden states contain recoverable uncertainty information, as shown by strong probe AUROC across layers. Third, latent uncertainty outperforms answer-only, scalar-score, and textual-uncertainty conditions as a downstream carrier.

These findings align with the main argument of the paper. When an upstream agent commits to an answer after calling a web-search tool, the committed answer does not fully preserve the uncertainty state that produced it. Scalar and textual annotations can recover part of the missing signal, but they remain limited. Latent uncertainty better preserves upstream fragility and therefore provides a stronger basis for downstream recovery behavior.


\section{Formalization of Interface Collapse}
\label{app:formalization}

This appendix provides a lightweight formalization of interface collapse. The goal is not to introduce a complete theory of agent uncertainty, but to clarify the structural claim made in the main paper: standard agent interfaces expose committed artifacts rather than uncertainty-bearing epistemic states. As a result, uncertainty may influence the upstream decision while failing to remain available for downstream recovery.

\subsection{Uncertainty-Bearing State}

Consider an upstream component that must produce an intermediate artifact for a downstream component. The artifact may be a tool call, a search query, a retrieved passage set, a structured field, an intermediate answer, or a final answer. Before committing to this artifact, the upstream component has an uncertainty-bearing state.

We represent this state as a distribution over possible commitments:
\[
\pi \in \Delta(\mathcal{Y}),
\]
where $\mathcal{Y}$ is the space of possible artifacts and $\Delta(\mathcal{Y})$ denotes the probability simplex over $\mathcal{Y}$. For example, in a web-search tool-use setting, $\mathcal{Y}$ may be the space of possible search queries. A sharp distribution indicates that the upstream component strongly prefers one query, while a diffuse distribution indicates that several queries remain plausible.

The important point is that $\pi$ contains more information than the final selected artifact. It can encode competing alternatives, confidence gaps, ambiguity, evidence conflict, or instability in the upstream decision. This is the information that may be useful for downstream recovery.

\subsection{Interface as a Decoding Map}

A standard agent interface requires the upstream component to emit a committed artifact. We formalize this interface as a decoding map:
\[
\Phi: \Delta(\mathcal{Y}) \rightarrow \mathcal{Y}.
\]
Given an uncertainty-bearing state $\pi$, the interface produces a single artifact:
\[
y^\ast = \Phi(\pi).
\]
For instance, $\Phi$ may select the most likely tool call, the highest-scoring search query, the top retrieved passage set, or the final answer selected by decoding.

This map is useful because downstream components need concrete artifacts to consume. A tool executor needs a tool name and arguments. A search API needs a query string. A parser needs structured fields. A verifier needs a claim or answer to inspect. However, this usefulness comes with an epistemic cost: $\Phi$ maps a distributional state into one committed object.

\subsection{Many-to-One Projection}

The central structural property is that $\Phi$ is generally many-to-one. There can exist two different uncertainty-bearing states $\pi_1$ and $\pi_2$ such that
\[
\pi_1 \neq \pi_2
\quad \text{but} \quad
\Phi(\pi_1) = \Phi(\pi_2) = y^\ast.
\]
This means that two upstream states with different uncertainty profiles can produce the same downstream artifact. A confident decision and a fragile decision may therefore become indistinguishable once they cross the interface.

For example, suppose an agent must choose a search query. One state may assign high probability to a single query:
\[
\pi_1(q_1)=0.95,\quad \pi_1(q_2)=0.03,\quad \pi_1(q_3)=0.02.
\]
Another state may be much more uncertain:
\[
\pi_2(q_1)=0.36,\quad \pi_2(q_2)=0.34,\quad \pi_2(q_3)=0.30.
\]
If the interface selects the top query, both states produce the same artifact:
\[
\Phi(\pi_1)=\Phi(\pi_2)=q_1.
\]
The downstream search tool receives the same query $q_1$ in both cases, but the epistemic status of that query is very different. In the first case, $q_1$ is a stable decision. In the second case, $q_1$ is only a narrow winner among competing alternatives.

\subsection{Interface Collapse}

We define interface collapse as the loss of distinguishability induced by the interface map. Before the interface, the upstream states $\pi_1$ and $\pi_2$ are distinct. After the interface, the downstream component observes only $y^\ast$ and cannot recover which uncertainty-bearing state produced it.

\begin{tcolorbox}[
  colback=gray!6,
  colframe=gray!55!black,
  title=\textbf{Interface Collapse},
  fonttitle=\bfseries,
  boxrule=0.8pt,
  arc=2mm,
  left=1.5mm,
  right=1.5mm,
  top=1mm,
  bottom=1mm
]
Interface collapse occurs when distinct uncertainty-bearing states are mapped to the same committed artifact:
\[
\exists \pi_1,\pi_2 \in \Delta(\mathcal{Y}) :
\pi_1 \neq \pi_2
\quad \text{and} \quad
\Phi(\pi_1)=\Phi(\pi_2).
\]
After this projection, downstream components receive the selected artifact but not the uncertainty profile behind it.
\end{tcolorbox}

This formalization captures why uncertainty does not automatically propagate through a trajectory. The trajectory may contain uncertain decisions, but uncertainty only remains operationally available if some uncertainty-bearing information survives the handoff. If the interface exposes only $y^\ast$, then the downstream component inherits the consequence of uncertainty without inheriting uncertainty itself.

\subsection{Collapse Magnitude}

One way to quantify the amount of information removed by the interface is to compare the entropy of the upstream distribution with the entropy of the committed artifact. The uncertainty-bearing state has entropy
\[
H(\pi) = - \sum_{y \in \mathcal{Y}} \pi(y)\log \pi(y).
\]
After deterministic commitment, the interface exposes a point mass on the selected artifact:
\[
\delta_{y^\ast}.
\]
The entropy of this committed artifact is
\[
H(\delta_{y^\ast}) = 0.
\]
A simple collapse magnitude can therefore be written as
\[
\mathcal{C}(\pi, \Phi)
=
H(\pi) - H(\delta_{\Phi(\pi)})
=
H(\pi).
\]
This quantity is not meant to capture all forms of epistemic loss. It only illustrates the basic asymmetry: before the interface, the upstream state may contain graded uncertainty over alternatives; after the interface, the standard artifact is treated as a single selected commitment.

\subsection{Why Re-estimation Is Not Propagation}

The formalization also clarifies the difference between uncertainty propagation and uncertainty re-estimation. Suppose the downstream component receives only $y^\ast$. It may still estimate a new uncertainty score:
\[
\hat{u}_{\mathrm{down}} = g(y^\ast),
\]
where $g$ is a downstream uncertainty estimator. This can be useful, but it is not equivalent to receiving the upstream state $\pi$.

The reason is that $g(y^\ast)$ is computed after the collapse. Since multiple upstream states may produce the same $y^\ast$, any function of $y^\ast$ alone must assign the same value to all such states:
\[
\Phi(\pi_1)=\Phi(\pi_2)=y^\ast
\quad \Rightarrow \quad
g(\Phi(\pi_1)) = g(\Phi(\pi_2)).
\]
Thus, downstream re-estimation cannot distinguish a confident upstream commitment from a fragile one when both produce the same artifact. It may detect risk from the artifact itself, but it cannot fully recover the uncertainty profile that was erased at the interface.

\subsection{Uncertainty-Preserving Handoff}

An uncertainty-preserving interface augments the committed artifact with an uncertainty carrier. We write such a handoff as
\[
\Psi(\pi) = \left(y^\ast, z_u\right),
\]
where
\[
y^\ast = \Phi(\pi)
\]
is the committed artifact and $z_u$ is an uncertainty-bearing carrier derived from the upstream state. The carrier may be a scalar confidence score, a textual uncertainty statement, a structured uncertainty schema, or a latent representation.

The purpose of $z_u$ is not necessarily to reconstruct the full upstream state $\pi$. Rather, it should preserve enough uncertainty-relevant information for downstream control. A downstream component can then condition its behavior on both the artifact and the uncertainty carrier:
\[
a_{\mathrm{down}} = h(y^\ast, z_u),
\]
where $a_{\mathrm{down}}$ may be a decision to continue, verify, retry, clarify, abstain, or reroute.

This view also explains the carrier comparison in Appendix~\ref{app:additional-results}. Different carriers preserve different amounts and kinds of information about $\pi$. An answer-only handoff sets
\[
\Psi_{\mathrm{answer}}(\pi)=y^\ast,
\]
which provides no explicit carrier. A scalar-score handoff provides a compressed risk estimate. A textual-uncertainty handoff provides a natural-language description of selected doubts. A latent-uncertainty handoff provides a representation-level carrier that may preserve uncertainty features not easily expressed as a single score or short text.

\subsection{Connection to Confidence Laundering}

Confidence laundering follows from interface collapse when the committed artifact has a form that appears more reliable than the upstream state warrants. Let $\pi$ be a fragile upstream state with high entropy or competing alternatives. If the interface emits only $y^\ast$, then downstream components may treat $y^\ast$ as procedurally valid:
\[
\text{fragile state } \pi
\quad \xrightarrow{\Phi}
\quad
\text{clean artifact } y^\ast.
\]
The artifact may be a valid query, a well-formed API call, a fluent answer, or a structured JSON field. Its format suggests readiness for use, even though the decision behind it may have been unstable.

Confidence laundering is therefore not merely overconfidence in the psychological sense. It is an interface-induced failure mode. The interface gives a fragile decision a clean operational form, and downstream components may over-trust that form because the fragility has been removed from the handoff.

\subsection{Summary}

This formalization supports three claims. First, standard agent interfaces can be modeled as maps from uncertainty-bearing states to committed artifacts. Second, these maps are generally many-to-one, so different uncertainty profiles can become indistinguishable after the handoff. Third, downstream uncertainty re-estimation is not the same as uncertainty propagation, because it operates after the interface has already collapsed the upstream state.

The practical implication is that agent systems should not only estimate uncertainty at individual steps. They should also decide what uncertainty-bearing information crosses the interface. An uncertainty-preserving handoff augments the committed artifact with a carrier that allows downstream components to act on fragility before it becomes cascading failure.

\section{Additional Results}
\label{app:additional-results}

This appendix provides additional empirical results for the web-search tool-use experiments described in Appendix~\ref{app:experimental-details}. The goal is not to introduce a separate experimental setting, but to report more complete measurements behind the main empirical observations. We focus on two questions. First, do different handoff carriers preserve different amounts of upstream uncertainty? Second, can the observed uncertainty be explained by HotpotQA difficulty labels alone?

These two analyses support the main argument of the paper. If uncertainty were merely a proxy for dataset difficulty, then external difficulty labels would be sufficient to explain the uncertainty behavior. If a scalar confidence score or textual uncertainty statement were sufficient to preserve upstream fragility, then a latent carrier would not provide additional value. The results in this appendix are designed to test these two possibilities.

\subsection{Carrier Comparison}
\label{app:carrier-comparison}

We first compare different handoff carriers between the upstream tool-use agent and the downstream uncertainty predictor. Each condition exposes a different amount or form of information after the upstream agent has answered a HotpotQA question using the Tavily web-search tool.

The answer-only condition corresponds to the standard committed interface. The downstream component receives only the final answer produced after web search. The scalar-score condition adds a compact confidence-like uncertainty signal. The textual-uncertainty condition adds a natural-language uncertainty description. The latent-uncertainty condition passes a representation-level carrier derived from the upstream hidden state.

Table~\ref{tab:carrier_comparison} reports AUROC and accuracy for each handoff condition. These metrics evaluate whether the downstream component can recover whether the upstream tool-mediated decision was fragile.

\begin{table}[t]                                                                                      
  \centering                                                                                            
  \small                                                                                                
  \renewcommand{\arraystretch}{1.15}                                                                    
  \setlength{\tabcolsep}{6pt}                                                                           
  \begin{tabular}{p{2.7cm}p{5.9cm}cc}
  \toprule                                                                                              
  \textbf{Handoff condition} & \textbf{Signal passed downstream} & \textbf{AUROC} & \textbf{Accuracy} \\
  \midrule                                                                                              
  Answer-only     
  & Final answer only                                                                                   
  & 0.618 & 0.703 \\ 
  Scalar score                                                                                          
  & Final answer + scalar uncertainty score
  & 0.667 & 0.767 \\                                                                                    
  Textual uncertainty
  & Final answer + textual uncertainty description                                                      
  & 0.785 & 0.792 \\      
  Latent uncertainty
  & Final answer + hidden-state carrier                                                                 
  & \textbf{0.874} & \textbf{0.893} \\                                                                  
  \bottomrule
  \end{tabular}                                                                                         
  \caption{Comparison of downstream uncertainty recoverability across different handoff conditions.
  Higher AUROC indicates that the handoff signal better preserves the upstream uncertainty behind the   
  tool-mediated decision.}
  \label{tab:carrier_comparison}                                                                        
  \end{table}      

\paragraph{Interpretation.}
This comparison is the most direct empirical test of our carrier argument. The answer-only condition measures how much uncertainty can be recovered from the committed artifact alone. The scalar-score and textual-uncertainty conditions test whether explicit but compressed uncertainty annotations are sufficient. The latent-uncertainty condition tests whether a representation-level carrier preserves additional uncertainty-relevant information.

If latent uncertainty outperforms the other handoff conditions, the result supports the interface-bottleneck view: the committed final answer does not fully preserve the upstream uncertainty that shaped the tool-use trajectory. Scalar and textual annotations may recover part of the missing signal, but they can still compress away information about the source and structure of fragility. The result should therefore be interpreted as evidence for stronger uncertainty recoverability, not as a claim that latent uncertainty is always the best carrier in every possible agent setting.

\subsection{Difficulty Labels versus Model Uncertainty}
\label{app:difficulty-vs-uncertainty}

We next examine whether the uncertainty labels used in our experiments can be explained by HotpotQA difficulty labels. HotpotQA marks questions as easy, medium, or hard, but these labels describe dataset-level question difficulty rather than the uncertainty of a particular model interacting with a web-search tool.

This distinction is important. A question labeled as hard may become stable if the Tavily search results contain direct and consistent evidence. Conversely, a question labeled as easy may induce uncertainty if the generated search query is ambiguous, if the returned snippets are incomplete, or if the model is sensitive to different pieces of web evidence. Therefore, model uncertainty in our setting should be understood as tool-use-specific fragility, not simply as external task difficulty.

Table~\ref{tab:difficulty_uncertainty} reports the relationship between HotpotQA difficulty labels and semantic-entropy-based uncertainty. For each difficulty group, we report the number of questions, the average number of semantic answer clusters, the mean semantic entropy, and the proportion of examples labeled as high uncertainty.

  \begin{table}[t]                                                                                      
  \centering
  \small                                                                                                
  \renewcommand{\arraystretch}{1.15}
  \setlength{\tabcolsep}{6pt}                                                                           
  \begin{tabular}{lcccc}
  \toprule                                                                                              
  \textbf{HotpotQA difficulty}
  & \textbf{\# Questions}                                                                               
  & \textbf{Mean cluster count}                                                                         
  & \textbf{Mean $H_{\mathrm{sem}}$}
  & \textbf{High-uncertainty rate} \\                                                                   
  \midrule        
  Easy   & 200 & 8.26 & 0.704 & 0.780 \\                                                                
  Medium & 200 & 8.19 & 0.759 & 0.820 \\                                                                
  Hard   & 200 & 7.97 & 0.706 & 0.770 \\
  \bottomrule                                                                                           
  \end{tabular}   
  \caption{Relationship between HotpotQA difficulty labels and model uncertainty under the Tavily       
  web-search tool-use setting. $H_{\mathrm{sem}}$ is the length-normalized semantic entropy (divided by 
  $\log K$); high-uncertainty rate uses the strict threshold ($H_{\mathrm{sem}} \geq 0.5$ \emph{and}
  largest-cluster ratio $\leq 0.5$). Difficulty labels are used only for analysis and are not used as   
  supervision for the uncertainty classifier.}
  \label{tab:difficulty_uncertainty}
  \end{table}              

\paragraph{Interpretation.}
This analysis tests whether uncertainty in our setting is reducible to external difficulty labels. If easy, medium, and hard questions show substantial overlap in cluster count, semantic entropy, or high-uncertainty rate, then difficulty labels alone are insufficient to explain the observed uncertainty behavior. This would support our claim that uncertainty is model-specific and interface-dependent. In the web-search setting, uncertainty can arise from query formulation, tool-output coverage, source consistency, entity resolution, and final answer generation, rather than only from the nominal difficulty of the original HotpotQA question.

\subsection{Summary of Additional Findings}
\label{app:additional-summary}

The additional results are intended to support two conclusions. First, different handoff carriers preserve different amounts of upstream uncertainty. If the latent-uncertainty condition provides stronger downstream uncertainty recoverability than answer-only, scalar-score, and textual-uncertainty conditions, then committed tool-use artifacts and compressed annotations do not fully preserve the fragility behind upstream decisions.

Second, the uncertainty observed in the Tavily web-search setting is not reducible to HotpotQA difficulty labels alone. Since difficulty labels are external dataset annotations, they cannot capture all uncertainty introduced by search-query formulation, web evidence coverage, evidence conflict, and model-specific answer instability. Together, these findings support the main argument that uncertainty propagation in agent systems should be treated as an interface problem: downstream components can only act on uncertainty if the handoff preserves it in a usable form.

\section{Examples of Uncertain Decision Handoffs}
\label{app:handoff-examples}

This appendix provides concrete examples of uncertain decision handoffs in the web-search tool-use setting described in Appendix~\ref{app:experimental-details}. The goal is not to introduce additional experiments, but to make the interface-bottleneck argument more concrete. In each case, an upstream agent makes an intermediate decision under uncertainty, exposes a clean artifact to the next component, and may lose part of the uncertainty that shaped the decision.

We focus on four common handoffs in a tool-mediated question-answering pipeline: search-query formulation, tool-result interpretation, evidence selection, and final-answer generation. These examples illustrate why uncertainty propagation cannot be reduced to final-answer confidence alone. The fragility may originate before the final answer is produced, but the downstream component may only receive the final committed artifact.

\subsection{Search-Query Handoff}
\label{app:search-query-handoff}

The first handoff occurs when the agent transforms a user question into a search query. In the Tavily-based setting, this query is the input to the external web-search tool. It is therefore not merely a natural-language reformulation; it is an operational artifact that determines what evidence the agent will receive.

\begin{tcolorbox}[
  colback=blue!4,
  colframe=blue!55!black,
  title=\textbf{Case D.1: Search-Query Handoff},
  fonttitle=\bfseries,
  boxrule=0.8pt,
  arc=2mm,
  left=1.5mm,
  right=1.5mm,
  top=1mm,
  bottom=1mm
]
\textbf{Question.}

Which actor appeared in the film directed by the director of \textit{Inception} and also played Jack Sparrow?

\medskip
\textbf{Possible internal uncertainty.}

The agent may be uncertain about which part of the question should be searched first. It may consider several plausible search queries:
\begin{itemize}
    \item \texttt{director of Inception filmography actors}
    \item \texttt{actor played Jack Sparrow film directed by Christopher Nolan}
    \item \texttt{Christopher Nolan film Johnny Depp}
\end{itemize}

\medskip
\textbf{Committed artifact.}

The standard interface exposes only one selected query, such as:
\[
\texttt{Christopher Nolan film Johnny Depp}
\]

\medskip
\textbf{Uncertainty lost at the handoff.}

The downstream tool call receives a clean query string, but it does not receive the fact that the agent was uncertain about the decomposition strategy, the entity relation, or the correct search direction. A weak query may still look well-formed, causing the next stage to treat it as a reliable starting point.
\end{tcolorbox}

Case D.1 illustrates an important feature of tool-use uncertainty. A search query can be syntactically valid while being epistemically fragile. Once the query is emitted, the tool interface preserves the decision but not the uncertainty behind it. The downstream observation is therefore conditioned on a choice whose fragility may no longer be visible.

\subsection{Tool-Result Handoff}
\label{app:tool-result-handoff}

The second handoff occurs when the external tool returns results to the agent or to a downstream component. In a standard web-search interface, Tavily returns titles, snippets, and URLs. These results are useful, but they are not equivalent to complete evidence. They may be partial, noisy, redundant, or indirectly relevant.

Case D.2 shows that tool results can hide uncertainty even when they are relevant. The problem is not simply retrieval failure. The issue is that partial evidence can be transformed into a clean intermediate representation, making downstream reasoning appear more certain than the underlying observation supports.

\subsection{Evidence-Selection Handoff}
\label{app:evidence-selection-handoff}

The third handoff occurs when the agent selects which pieces of evidence to use. Web search often returns several snippets that vary in relevance and specificity. The agent must decide which snippets support the answer and which should be ignored. This selection step is uncertainty-sensitive because the final answer may depend heavily on which evidence is treated as authoritative.

Case D.3 illustrates why uncertainty can be source-structured rather than scalar. The risk is not only that the agent has low confidence. The risk is that the agent is uncertain about the type of comparison, the entity boundary, or the evidence unit. A single confidence score may indicate that the answer is risky, but it may not preserve the structure needed to choose the right recovery action.
\begin{tcolorbox}[
  colback=yellow!8,
  colframe=yellow!45!black,
  title=\textbf{Case D.2: Tool-Result Handoff},
  fonttitle=\bfseries,
  boxrule=0.8pt,
  arc=2mm,
  left=1.5mm,
  right=1.5mm,
  top=1mm,
  bottom=1mm
]
\textbf{Question.}

What country was the composer of the soundtrack for \textit{The Dark Knight} born in?

\medskip
\textbf{Search query.}

\[
\texttt{The Dark Knight soundtrack composer birthplace}
\]

\medskip
\textbf{Possible Tavily observation.}

The returned snippets may mention that the soundtrack was composed by Hans Zimmer and James Newton Howard. Some snippets may mention Hans Zimmer's birthplace, while others may focus on James Newton Howard or on the film soundtrack more generally.

\medskip
\textbf{Possible internal uncertainty.}

The agent may be uncertain about which composer the question refers to, whether the soundtrack should be attributed to one composer or both, and whether the returned snippets are sufficient to answer the question.

\medskip
\textbf{Committed artifact.}

The downstream component may receive a compact observation such as:
\[
\texttt{Hans Zimmer composed the soundtrack. Hans Zimmer was born in Germany.}
\]

\medskip
\textbf{Uncertainty lost at the handoff.}

The downstream component receives a clean evidence summary, but it may not receive the unresolved ambiguity that there were multiple composers. The apparent clarity of the summary can launder uncertainty into confidence.
\end{tcolorbox}

\begin{tcolorbox}[
  colback=green!5,
  colframe=green!45!black,
  title=\textbf{Case D.3: Evidence-Selection Handoff},
  fonttitle=\bfseries,
  boxrule=0.8pt,
  arc=2mm,
  left=1.5mm,
  right=1.5mm,
  top=1mm,
  bottom=1mm
]
\textbf{Question.}

Which university is older, the university attended by the author of \textit{The Great Gatsby}, or the university where the inventor of the World Wide Web studied?

\medskip
\textbf{Possible evidence returned by web search.}

The search results may mention that F. Scott Fitzgerald attended Princeton University and that Tim Berners-Lee studied at The Queen's College, Oxford. Other snippets may mention Oxford University more generally, or may describe Fitzgerald's short attendance at Princeton without graduation.

\medskip
\textbf{Possible internal uncertainty.}

The agent may be uncertain about whether to compare Princeton University with The Queen's College, Oxford, or with the University of Oxford as an institution. It may also be uncertain whether ``attended by'' requires graduation or enrollment.

\medskip
\textbf{Committed artifact.}

The agent may pass forward a selected evidence pair:
\[
\texttt{Fitzgerald attended Princeton; Berners-Lee studied at Oxford.}
\]

\medskip
\textbf{Uncertainty lost at the handoff.}

The selected evidence pair looks complete, but it hides the entity-resolution decision. The downstream component may compare founding years without realizing that the upstream component made an uncertain choice about the institutional unit being compared.
\end{tcolorbox}

\subsection{Final-Answer Handoff}
\label{app:final-answer-handoff}

The fourth handoff occurs when the agent emits the final answer. This is the most visible interface, and it is often the only artifact consumed by downstream components. However, by the time the final answer is emitted, uncertainty from earlier stages may already have been compressed away.

\begin{tcolorbox}[
  colback=red!4,
  colframe=red!50!black,
  title=\textbf{Case D.4: Final-Answer Handoff},
  fonttitle=\bfseries,
  boxrule=0.8pt,
  arc=2mm,
  left=1.5mm,
  right=1.5mm,
  top=1mm,
  bottom=1mm
]
\textbf{Question.}

Was the person who founded the company that created the iPhone born before the founder of Microsoft?

\medskip
\textbf{Tool-mediated reasoning path.}

The agent may search for the founder of Apple, identify Steve Jobs, search for the founder of Microsoft, identify Bill Gates, and compare their birth years.

\medskip
\textbf{Possible internal uncertainty.}

The agent may be uncertain because Apple had multiple founders, because the question refers to the company that created the iPhone rather than directly naming Apple, or because the answer depends on which founder is selected.

\medskip
\textbf{Committed final answer.}

\[
\texttt{Yes. Steve Jobs was born in 1955, before Bill Gates, who was born in 1955.}
\]

\medskip
\textbf{Uncertainty lost at the handoff.}

The final answer has the form of a confident comparison, but it may hide unresolved ambiguity about founder selection and exact date comparison. A downstream verifier receives a polished answer, not the upstream uncertainty that should have triggered a clarification or a more careful comparison.
\end{tcolorbox}

The final-answer handoff is where confidence laundering becomes most visible. A fluent answer can appear stable even when the upstream trajectory contained unstable entity choices, weak evidence, or incomplete comparison logic. This is why answer-only handoffs are insufficient for studying uncertainty propagation in agent systems.

\subsection{Prompt-Level Illustration}
\label{app:prompt-level-illustration}

For completeness, we provide an illustrative prompt template used to construct the web-search tool-use trajectory. This template is not intended as a contribution by itself. It is included to clarify how a HotpotQA question is converted into a tool-mediated decision process.

\begin{tcolorbox}[
  colback=purple!4,
  colframe=purple!55!black,
  title=\textbf{Case D.5: Prompt Template for Tool Use},
  fonttitle=\bfseries,
  boxrule=0.8pt,
  arc=2mm,
  left=1.5mm,
  right=1.5mm,
  top=1mm,
  bottom=1mm
]
\textbf{System instruction.}

You are a question-answering agent with access to a web-search tool. Given a question, first produce a search query that would help answer the question. After receiving the web-search results, answer the question using only the provided evidence. If the evidence is incomplete or conflicting, give the best supported answer and do not invent unsupported facts.

\medskip
\textbf{User input.}

\begin{verbatim}
Question: {hotpotqa_question}
\end{verbatim}

\textbf{Tool call.}

\begin{verbatim}
Search query: {model_generated_query}
\end{verbatim}

\textbf{Tool observation.}

\begin{verbatim}
Tavily results: {title, snippet, url}_{1:n}
\end{verbatim}

\textbf{Final response.}

\begin{verbatim}
Answer: {final_answer}
\end{verbatim}
\end{tcolorbox}

This prompt creates a simple but explicit tool-use interface. The question is not answered from a fixed retrieved context. Instead, the model must pass through a search-query decision, a Tavily tool observation, an evidence-use step, and a final answer. Each stage introduces a possible uncertain decision handoff.

\subsection{Summary}
\label{app:handoff-examples-summary}

These examples show that uncertainty in tool-use agents can originate at multiple interfaces. A search query can hide uncertainty about the intended information need. A tool observation can hide uncertainty about evidence coverage. A selected evidence pair can hide uncertainty about entity resolution or comparison scope. A final answer can hide uncertainty accumulated throughout the trajectory.

The key point is that downstream components often receive clean artifacts rather than uncertainty-bearing states. This is the mechanism behind interface collapse. The artifact remains usable, but the fragility behind it may no longer be accessible. These examples therefore motivate the carrier comparison in Appendix~\ref{app:additional-results}: answer-only, scalar-score, textual-uncertainty, and latent-uncertainty handoffs differ in how much upstream fragility they preserve for downstream recovery.

\newpage

\end{document}